\documentclass[conference]{IEEEtran}
\IEEEoverridecommandlockouts
\usepackage{cite}
\usepackage{amsmath,amssymb,amsfonts}
\usepackage{algorithmic}
\usepackage{graphicx}
\usepackage{textcomp}
\PassOptionsToPackage{hyphens}{url}
\usepackage{hyperref} 
\usepackage[dvipsnames]{xcolor}
\usepackage{listings,multicol}

\makeatletter
\def\lst@makecaption{%
  \def\@captype{table}%
  \@makecaption
}
\makeatother
\def\BibTeX{{\rm B\kern-.05em{\sc i\kern-.025em b}\kern-.08em
    T\kern-.1667em\lower.7ex\hbox{E}\kern-.125emX}}

\usepackage{dirtytalk} 
\usepackage{multirow} 
\usepackage{subcaption}

\usepackage[utf8]{inputenc} 
\usepackage[T1]{fontenc}

\usepackage[textsize=tiny, textwidth=29]{todonotes}

\usepackage{tabularx}

\begin{document}

\title{The Importance and the Limitations of Sim2Real for Robotic Manipulation in Precision Agriculture}
\author{Carlo Rizzardo, Sunny Katyara, Miguel Fernandes, Fei Chen%
\thanks{This research is supported in part by the project ``Grape Vine Perception and Winter Pruning Automation'' funded by joint lab of Istituto Italiano di Tecnologia and Università Cattolica del Sacro Cuore, and the project ``Improving Reproducibility in Learning Physical Manipulation Skills with Simulators Using Realistic Variations'' funded by EU H2020 ERA-Net Chist-Era program. \textit{(Corresponding author: Fei Chen)} }
\thanks{ Carlo Rizzardo, Sunny Katyara, Miguel Fernandes and Fei Chen are with Active Perception and Robot Interactive Learning Laboratory, Department of Advanced Robotics, Istituto Italiano di Tecnologia, Via Morego 30, 16163, Genova, Italy.}
}

\markboth{IIT - Active Perception And Robot Interactive Learning Laboratory}%
{IIT - Active Perception And Robot Interactive Learning Laboratory}

\maketitle

\begin{abstract}
In recent years Sim2Real approaches have brought great results to robotics. Techniques such as model-based learning or domain randomization can help overcome the gap between simulation and reality, but in some situations simulation accuracy is still needed. An example is agricultural robotics, which needs detailed simulations, both in terms of dynamics and visuals. However, simulation software is still not capable of such quality and accuracy. Current Sim2Real techniques are helpful in mitigating the problem, but for these specific tasks they are not enough.
\end{abstract}



\section{Introduction} \label{intro}
Sim2Real methods bring significant advantages to robotics.
They allow to test and train robots to perform tasks without having direct access to the physical environment. 
This accelerates the development by providing a readily available and reproducible environment.
In addition to this, the capability of simulating physics at rates greater than reality can accelerate testing and training when using learning approaches.

These advantages are particularly helpful to agricultural and Agri-food robotics.
Compared to industrial applications, agricultural tasks are subject to additional constraints that can restrict the use of the real environment. 
In particular, agriculture is bound to the evolution of seasons, and testing of particular tasks is limited to extremely narrow time ranges within the year. 
Also, performing tasks such as harvesting \cite{zhao2016review} and pruning \cite{botterill2017robot} changes the environment irreversibly.
This implies the necessity of a great number of plants to experiment on, which are often not available, as they are valuable to the agricultural production and cannot risk to be damaged. 

However, simulating agricultural robotics setups is extremely challenging in terms of visual rendering and physics modeling. 
Natural elements can have specific texture and reflective characteristics that are problematic to simulate accurately.
Also, plants, branches, and fruits have deformation dynamics which are challenging to simulate and design with sufficient accuracy to support robot manipulation skills. 
To complicate the situation further, natural elements are extremely heterogeneous. To fully cover the range of variations among different specimens it is necessary  to design a great number of models, or devise methods to generate them programmatically. 

\section{The Simulator} \label{sec:simChoice}
Using simulation-based approaches in agriculture robotics without introducing sim-to-real transfer issues is particularly tricky.
The most crucial factors to take into account selecting the software are: rendering quality, physics simulation accuracy and ease of integration with other software (e.g., ROS). 

\textit{Gazebo} is usually the first choice for projects based on ROS as it provides a wide array of functionalities through the ROS messaging system, including control of the simulation execution, simulated controllers, and access to simulated sensors. 

Regarding physics simulation, \textit{Gazebo} supports four different engines: \textit{ODE} \cite{ode}, \textit{Bullet Physics} \cite{bullet}, \textit{DART} \cite{dart}, and \textit{Simbody} \cite{simbody}. 
This wide choice guarantees robustness and flexibility, as it is possible to select the most suited engine for each of the specific tasks in a project. 
\textit{Bullet Physics} in particular achieves state-of-the-art performance and accuracy, reason for which it is commonly used in Reinforcement Learning research \cite{peng2018deepmimic}\cite{kalashnikov2018qt}. 
However, due to its now old integration within \textit{Gazebo},  newer features are not exploited (the Featherstone solver is not being used). 

\textit{Gazebo} uses the \textit{OGRE} rendering engine, its capabilities are not on the level of state-of-the-art photorealistic engines such as \textit{Unreal Engine}, \textit{Unity} or \textit{Nvidia Omniverse}. 
Nonetheless, it is still able to provide quite realistic renderings \cite{allan2019planetary}.

As mentioned, some components of \textit{Gazebo} are starting to show the age of its codebase.
A rewrite of the software is undergoing, named \textit{Gazebo Ignition}.
It promises to improve rendering capabilities by using \textit{OGRE v2} instead of \textit{OGRE v1}, and also physics simulation quality will be enhanced with the re-integration of \textit{Bullet Physics} \cite{ignGazeboBullet} and \textit{DART} \cite{ignGazeboComparison}. 

Potential alternatives to \textit{Gazebo} are \textit{Webots} \cite{webots}, \textit{CoppeliaSim} \cite{coppeliaSim} (previously \textit{V-REP}) or \textit{Unity}.
However, while they may be better in specific aspects, \textit{Gazebo} is still overall more suited to the requirements of agricultural robotics. 
For example \textit{Unity} would provide higher quality rendering, but at the expense of a less straightforward ROS integration. 
\textit{Webots} is on par with \textit{Gazebo} for what concerns the rendering, but it does not offer the same flexibility in the physics simulation, and, also in this case, ROS integration is less obvious. 
\textit{CoppeliaSim} offers physics simulation capabilities comparable to those of \textit{Gazebo}, but the rendering is more limited. 

In future, other options will be viable. 
Most notably, \textit{Nvidia} is developing a simulator within its \textit{Isaac} framework, \textit{Nvidia Isaac Sim}.
Its visual rendering, based on \textit{Nvidia Omniverse} and ray tracing technology, will be of extremely high and photorealistic quality.
Also, the accuracy of its physics simulation, based on \textit{PhysX 5}, will be state of the art. 

\section{An Example: Robotic Grape Vine Pruning}
A task representative of the highlighted difficulties is the winter pruning of grapevines, which is the center of an undergoing project in our lab \cite{aprilVinum}. 
In wine-producing vineyards grapevines require to be pruned during the winter and trimmed in the summer to control the growth of the plant, the number and position of new shoots, and the number and quality of the grape clusters. 
It is a job that requires the work of numerous skilled laborers for a limited period every year, consequently the available workforce is scarce \cite{poni2016mechanical}. 
Because of this, it has great appeal for automation.
However, experiments in the real world are constrained by the short supply of vines and by the narrow window of time in which tests can be performed \cite{gatti2017ground}. 

To achieve an optimal balance between quality and quantity of the grapes, it is important to choose carefully the branches to be removed, based on criteria such as the width and length of the branches, their distance from the main cordon or the locations of the buds.
This analysis requires high-fidelity rendering and realistic models. 
Furthermore, pruning is a task rich of contact interactions between the robot tool and the vine, and the dynamics of the plant are particularly complex due to the numerous intertwined and flexible branches. 
Realistic simulation of these dynamics is crucial for learning and fine-tuning robot manipulation skills.
Even when not using learning approaches, just a slight movement due to collisions between robot and branches can lead to an incorrect cut if not taken into account.

\subsection{Sim2Real Setup for Grapevine Pruning}

In our implementation a robot composed of a wheeled mobile platform and a robot arm has to navigate in the vineyard, locate the vines, identify spurs and pruning locations, and proceed to cut the branches. 

A simulated setup was created in \textit{Gazebo} for testing the solution, with a simple environment but detailed models for the single vines.
The vine models were designed in \textit{Blender}, taking images of real vines as a guide.
To approximate the motion of the vines, the base of the plant was actuated randomly on three joints to induce translational and rotational motions.

The detection system for spurs and pruning regions has been implemented employing a \textit{Faster R-CNN} pipeline \cite{ren2015faster}.
The system was trained on a real-world dataset collected during the winter in a non-pruned vineyard. 
This solution was able to generalize to the simulation, allowing for simulated testing of the system.
The detection of the pruning points is performed analyzing close-up images of the spurs and extracting graph-morphometry information. The technique has been employed to generate cutting poses in both simulation and real-world.

\subsection{Limitations and potential solutions}
Even though working in simulation has led to interesting results in our project, the lack of realism in visual rendering and physical simulation is a huge limiting factor. 
Designing visually realistic vine models is extremely time-consuming, even in a context like our own, where leaves are absent.
And even a carefully designed model may not represent correctly minute characteristics such as young and small buds.
Even more challenging is the modeling of the dynamics of the plants, in particular their deformability.
The simulation of complex deformable objects is not yet a common and established feature in simulators.
Recent works tend to only treat simple and small objects \cite{wu2019learning}\cite{matas2018sim}.
A compromise is to insert joints in the plant structure, but, if not by using a considerable amount of joints, this does not lead to realistic results.
An approach like this can be extremely time-consuming at the model design stage.
Also, while approaches based on careful hand-crafting of the models can lead to sufficient realism for traditional methods based on optimal control or classical computer vision, the accuracy of such simulation may not be enough for learning-based methods.

\begin{figure}
	\centering
	\begin{subfigure}{.22\textwidth}
		\centering
		\includegraphics[width=.95\linewidth]{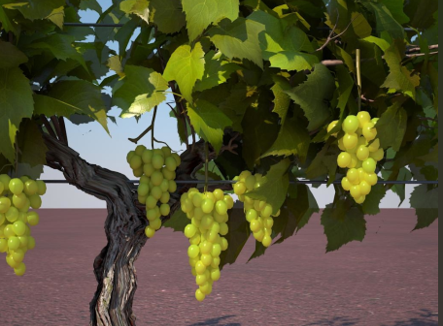}
		\caption{Vine model in 3ds Max}
		\label{fig:sub3}
	\end{subfigure}
	\begin{subfigure}{.22\textwidth}
		\centering
		\includegraphics[width=.95\linewidth]{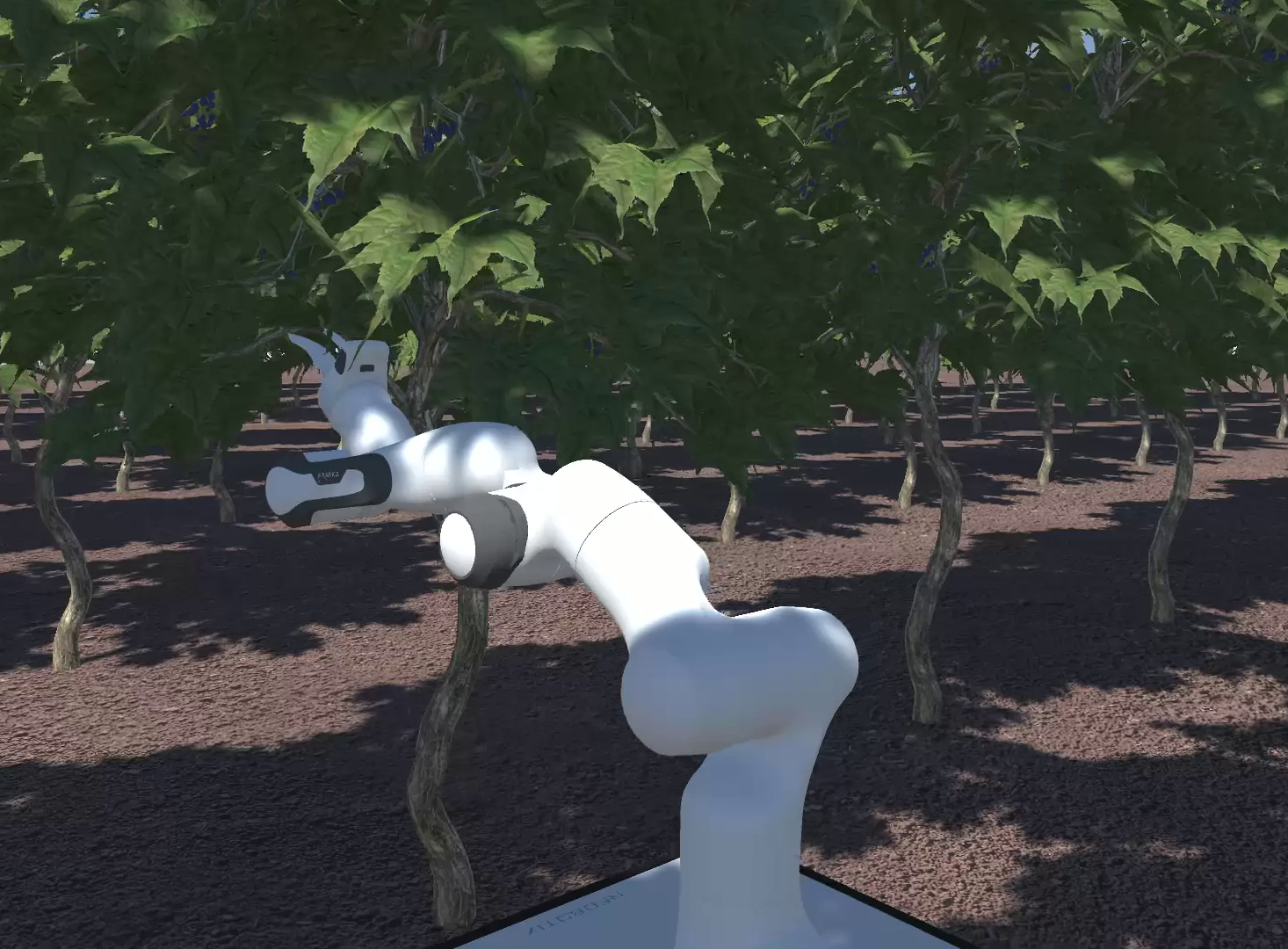}
		\caption{Vineyard and robot in Unity}
		\label{fig:sub4}
	\end{subfigure}
	\caption{Real and simulated grapevines}
	\label{fig:vines}
	\vspace{-10px}
\end{figure}

All of these issues are further complicated by the absence of a single simulation software capable of integrating state-of-the-art rendering and physics simulation, together with a good support for robotics applications.
\textit{Gazebo} is greatly integrated within the ROS framework, can provide accurate physics simulation and sufficiently good rendering.
However, creating visually realistic models for it can be challenging. 
Other software, like \textit{Unity}, can provide visual rendering of photorealistic quality, and recently there has been a push to include accurate physics simulation methods \cite{unityFeatherstone} but integration into robotics projects can be cumbersome. 
Still, preliminary trials in creating visually realistic environments show the potential of such software. Figure \ref{fig:vines} shows such a test in which a static vineyard environment was created.



\section{Conclusion and Discussion}

In the field of agricultural robotics simulation is a necessity. 
The seasonal constraints in this field, together with the scarcity and the monetary value of testing plants, do not allow extensive testing or training in the real world. 
High quality simulation of visuals and dynamics is needed, but plants have highly heterogeneous characteristics. 
Consequently there is a necessity for a great number of very complex models, or alternatively of procedurally generated ones. 
Currently, to our knowledge, there is no simulation software capable of satisfying all of the requirements of this area of research while also allowing straightforward integration with robotics frameworks.
There are promising projects like \textit{Gazebo Ignition}, \textit{Nvidia Isaac Sim} or \textit{Unity}, but they are not yet mature.
Agriculture robotics is an area of research that requires simulation, and simulation software is not ready yet to support it effectively without introducing sim-to-real transfer issues.


%

\bibliographystyle{IEEEtran}
\bibliography{sim2real}

\begin{thebibliography}{10}
\providecommand{\url}[1]{#1}
\csname url@samestyle\endcsname
\providecommand{\newblock}{\relax}
\providecommand{\bibinfo}[2]{#2}
\providecommand{\BIBentrySTDinterwordspacing}{\spaceskip=0pt\relax}
\providecommand{\BIBentryALTinterwordstretchfactor}{4}
\providecommand{\BIBentryALTinterwordspacing}{\spaceskip=\fontdimen2\font plus
\BIBentryALTinterwordstretchfactor\fontdimen3\font minus
  \fontdimen4\font\relax}
\providecommand{\BIBforeignlanguage}[2]{{%
\expandafter\ifx\csname l@#1\endcsname\relax
\typeout{** WARNING: IEEEtran.bst: No hyphenation pattern has been}%
\typeout{** loaded for the language `#1'. Using the pattern for}%
\typeout{** the default language instead.}%
\else
\language=\csname l@#1\endcsname
\fi
#2}}
\providecommand{\BIBdecl}{\relax}
\BIBdecl

\bibitem{zhao2016review}
Y.~Zhao, L.~Gong, Y.~Huang, and C.~Liu, ``A review of key techniques of
  vision-based control for harvesting robot,'' \emph{Computers and Electronics
  in Agriculture}, vol. 127, pp. 311--323, 2016.

\bibitem{botterill2017robot}
T.~Botterill, S.~Paulin, R.~Green, S.~Williams, J.~Lin, V.~Saxton, S.~Mills,
  X.~Chen, and S.~Corbett-Davies, ``A robot system for pruning grape vines,''
  \emph{Journal of Field Robotics}, vol.~34, no.~6, pp. 1100--1122, 2017.

\bibitem{ode}
\BIBentryALTinterwordspacing
``{ODE - Open Dynamics Engine}.'' [Online]. Available:
  \url{https://www.ode.org/}
\BIBentrySTDinterwordspacing

\bibitem{bullet}
\BIBentryALTinterwordspacing
``{Bullet Physics}.'' [Online]. Available:
  \url{https://pybullet.org/wordpress/}
\BIBentrySTDinterwordspacing

\bibitem{dart}
\BIBentryALTinterwordspacing
``{DART - Dynamic Animation and Robotics Toolkit}.'' [Online]. Available:
  \url{http://dartsim.github.io/}
\BIBentrySTDinterwordspacing

\bibitem{simbody}
\BIBentryALTinterwordspacing
``{Simbody: Multibody Physics API}.'' [Online]. Available:
  \url{https://simtk.org/projects/simbody/}
\BIBentrySTDinterwordspacing

\bibitem{peng2018deepmimic}
X.~B. Peng, P.~Abbeel, S.~Levine, and M.~van~de Panne, ``Deepmimic:
  Example-guided deep reinforcement learning of physics-based character
  skills,'' \emph{ACM Transactions on Graphics (TOG)}, vol.~37, no.~4, pp.
  1--14, 2018.

\bibitem{kalashnikov2018qt}
D.~Kalashnikov, A.~Irpan, P.~Pastor, J.~Ibarz, A.~Herzog, E.~Jang, D.~Quillen,
  E.~Holly, M.~Kalakrishnan, V.~Vanhoucke \emph{et~al.}, ``Qt-opt: Scalable
  deep reinforcement learning for vision-based robotic manipulation,''
  \emph{arXiv preprint arXiv:1806.10293}, 2018.

\bibitem{allan2019planetary}
M.~Allan, U.~Wong, P.~M. Furlong, A.~Rogg, S.~McMichael, T.~Welsh, I.~Chen,
  S.~Peters, B.~Gerkey, M.~Quigley \emph{et~al.}, ``Planetary rover simulation
  for lunar exploration missions,'' in \emph{2019 IEEE Aerospace
  Conference}.\hskip 1em plus 0.5em minus 0.4em\relax IEEE, 2019, pp. 1--19.

\bibitem{ignGazeboBullet}
\BIBentryALTinterwordspacing
``{Ignition Physics Issue 44 - Integrate Bullet}.'' [Online]. Available:
  \url{https://github.com/ignitionrobotics/ign-physics/issues/44}
\BIBentrySTDinterwordspacing

\bibitem{ignGazeboComparison}
\BIBentryALTinterwordspacing
``{Ignition Citadel - Feature comparison}.'' [Online]. Available:
  \url{https://www.ignitionrobotics.org/docs/citadel/comparison}
\BIBentrySTDinterwordspacing

\bibitem{webots}
\BIBentryALTinterwordspacing
``{Webots},'' accessed 2020-06-17. [Online]. Available:
  \url{https://www.cyberbotics.com/}
\BIBentrySTDinterwordspacing

\bibitem{coppeliaSim}
\BIBentryALTinterwordspacing
``{CoppeliaSim},'' accessed 2020-03-24. [Online]. Available:
  \url{https://www.coppeliarobotics.com/coppeliaSim}
\BIBentrySTDinterwordspacing

\bibitem{aprilVinum}
\BIBentryALTinterwordspacing
{Active Perception and Robot Interactive Learning Laboratory - Department of
  Advanced Robotics - Istituto Italiano di Tecnologia}, ``{Grape Vine Winter
  Pruning},'' accessed 2020-06-22. [Online]. Available:
  \url{https://advr.iit.it/research/april/grape-vine}
\BIBentrySTDinterwordspacing

\bibitem{poni2016mechanical}
S.~Poni, S.~Tombesi, A.~Palliotti, V.~Ughini, and M.~Gatti, ``Mechanical winter
  pruning of grapevine: physiological bases and applications,'' \emph{Scientia
  Horticulturae}, vol. 204, pp. 88--98, 2016.

\bibitem{gatti2017ground}
M.~Gatti, A.~Garavani, A.~Vercesi, and S.~Poni, ``Ground-truthing of remotely
  sensed within-field variability in a cv. barbera plot for improving vineyard
  management,'' \emph{Australian journal of grape and wine research}, vol.~23,
  no.~3, pp. 399--408, 2017.

\bibitem{ren2015faster}
S.~Ren, K.~He, R.~Girshick, and J.~Sun, ``Faster r-cnn: Towards real-time
  object detection with region proposal networks,'' in \emph{Advances in neural
  information processing systems}, 2015, pp. 91--99.

\bibitem{wu2019learning}
Y.~Wu, W.~Yan, T.~Kurutach, L.~Pinto, and P.~Abbeel, ``Learning to manipulate
  deformable objects without demonstrations,'' \emph{arXiv preprint
  arXiv:1910.13439}, 2019.

\bibitem{matas2018sim}
J.~Matas, S.~James, and A.~J. Davison, ``Sim-to-real reinforcement learning for
  deformable object manipulation,'' \emph{arXiv preprint arXiv:1806.07851},
  2018.

\bibitem{unityFeatherstone}
\BIBentryALTinterwordspacing
``{Unity - ArticulationBody}.'' [Online]. Available:
  \url{https://docs.unity3d.com/2020.2/Documentation/ScriptReference/ArticulationBody.html}
\BIBentrySTDinterwordspacing

\end{thebibliography}

\end{document}